\documentclass[
]{ceurart}

\usepackage{listings}
\begin{document}

\copyrightyear{2022}
\copyrightclause{Copyright for this paper by its authors.
  Use permitted under Creative Commons License Attribution 4.0
  International (CC BY 4.0).}

\conference{Forum for Information Retrieval Evaluation, December 9-13, 2022, India}
\title{Summarizing Indian Languages using Multilingual Transformers based Models}


\author[1]{Dhaval Taunk}[%
orcid=0000-0001-7144-4520,
email=dhaval.taunk@research.iiit.ac.in,
url=https://dhavaltaunk08.github.io//,
]
\address[1]{International Institute of Information Technology, Hyderabad, Telangana, India}

\author[1]{Vasudeva Varma}[%
email=vv@iiit.ac.in,
url=https://www.iiit.ac.in/~vv,
]



\begin{abstract}
  With the advent of multilingual models like mBART, mT5, IndicBART etc., summarization in low resource Indian languages is getting a lot of attention now a days. But still the number of datasets is low in number. In this work, we (Team HakunaMatata) study how these multilingual models perform on the datasets which have Indian languages as source and target text while performing summarization. We experimented with IndicBART and mT5 models to perform the experiments and report the ROUGE-1, ROUGE-2, ROUGE-3 and ROUGE-4 scores as a performance metric. 
\end{abstract}


\maketitle

\section{Introduction}

Automatic text summarization has a lot of potential applications in the current technological era like summarizing news articles, research articles etc. A lot of work has already been done in summarizing English languages text. But very little work is being done in summarizing Indian Languages. Therefore, summarizing text in these languages apart from English has become an essential task. India has approximately 350 million and 50 million Hindi and Gujarati speakers respectively. So building a summarization model in these languages will play a crucial role for this task. Recently, transformers based models like mBart\cite{liu-etal-2020-multilingual-denoising}, mT5\cite{xue-etal-2021-mt5} and IndicBart\cite{dabre-etal-2022-indicbart} have gained a lot of attention because of their multilingual capabilities including various Indic Languages.

Summarization can be performed in 2 ways: extractive summarization and abstractive summarization. In extractive summarization, a subset of sentences from the input text is taken as output summary. While in abstractive summarization, the entire summary is generated from scratch with the source text as input. Since text in abstractive summarization, summary is generated from scratch, this makes it more human like generated text. But at the same time, it becomes more difficult to perform abstractive summarization as compared to extractive summarization. 

In this work, we aim to perform abstractive summarization on these languages as a part of the FIRE shared task 2022 - ILSUM \cite{fire2022-ilsum}\cite{fire2022-ilsum-acm} using the dataset provided by the organizers. We used IndicBART and mT5 models for our experiments. We also performed data augmentation and tested the performance of the models. In the last, we report the ROUGE-1, ROUGE-2, ROUGE-3 and ROUGE-4 scores as mentioned by the shared task organizers.

\section{Related Work}

Both the extractive and abstractive summarization are well explored problem in English language context. A lot of datasets are available in English. Pubmed\cite{pubmed}, arXiv\cite{arxiv}, CNN/Daily Mail\cite{cnn} are to name a few. 

Guo et. al.,\cite{guo-etal-2022-longt5} extended T5\cite{2020t5} model to take long text as input and performed summarization over PubMed dataset. PRIMERA\cite{xiao-etal-2022-primera} is also another model which uses Longformer\cite{Beltagy2020Longformer} model and achieved state of the art results on datasets like arXiv summarization data\cite{arxiv}, Multi-News\cite{fabbri-etal-2019-multi} and WCEP\cite{gholipour-ghalandari-etal-2020-large} datasets. Hasan et. al.,\cite{hasan-etal-2021-xl} introduced a multilingual dataset named XL-Sum comprising of 44 languages. They experimented with mT5\cite{xue-etal-2021-mt5} model to perform abstractive summarization and report results based upon that.

Aries et. al.,\cite{aries-etal-2015-allsummarizer} performed multilingual and multi document summarization by clustering sentences into topics using a fuzzy clustering algorithm. They score each sentence based upon the topic coverage and then they create summary using the highest scoring sentences. For cross lingual abstractive summarization, Ladhak et. al.,\cite{ladhak-etal-2020-wikilingua} proposed WikiLingua, a article-summary pairs multilingual dataset available in 18 different languages. They fine-tuned mBART\cite{liu-etal-2020-multilingual-denoising} in their experiments.


\section{Methodology}

The main aim of the task is to generate summary for the articles and headline pairs 3 languages viz. English, Hindi, Gujarati. Although news articles and headlines have been used in a number of earlier efforts in other languages, the current dataset presents a special problem of code- and script-mixing. Even though the item is written in an Indian language, phrases from English are frequently used in news stories. We perform experiments using IndicBart and mT5 models after performing some data analysis. We also found data augmentation to be a useful approach in getting better results.

\subsection{Data Description}

The dataset provided by the organizers is divided into 3 languages with 3 splits (train, val and test) present for all these 3 languages. Table \ref{tab:count} shows the article count for all 3 splits across all the languages. 

For training phase, we were provided with the id of the article, link to the article, heading, summary and article text. While for the testing phase, we we given id of the article and the corresponding article text.

Since no references summaries was provided for the validation set. Therefore, while performing our own experiment, we took a small subset from the train set as our in-house validation set and then performed experiments over these sets. While we performed experiments on our in-house validation set, during the validation phase, we evaluated our model on the official validation set. After that, since there was limitation of 3 submission per languages, we chose our top 3 performing experiments of each language as our final submission for test phase.

\begin{table*}[h!]
  \caption{Number of instances per languages per split}
  \label{tab:count}
  \begin{tabular}{cccc}
    \toprule
    \textbf{Language} & \textbf{Train} & \textbf{Validation} & \textbf{Test} \\
    \midrule
    \multirow{1}{*}{English} & 12565 & 898 & 4487 \\
     \midrule
    \multirow{1}{*}{Hindi} & 7957 & 569 & 2842 \\
     \midrule
    \multirow{1}{*}{Gujarati} & 8457 & 605 & 3020 \\
  \bottomrule
\end{tabular}
\end{table*}

\section{Experiments}

This section explains the steps that we undergo to perform the experiments and also the different experiments that we performed on the dataset.

\subsection{Models used}

For our experiments, we fine-tuned models viz. IndicBART and mT5-small the details of which is given below:

\begin{enumerate}
    \item \textbf{IndicBART}: The eleven Indic languages and English are the main focus of the multilingual, sequence-to-sequence pre-trained model known as IndicBART. The authors tested IndicBART on two NLG tasks: extreme summarization and neural machine translation (NMT) and demonstrated that despite being substantially smaller, models IndicBART is competitive with huge pre-trained models like mBART50.
    \item \textbf{mT5}: A new Common Crawl-based dataset with 101 languages was used to pre-train the multilingual T5 model (mT5 model). For mT5, the model design and training process closely resemble those of T5.
\end{enumerate}

Both the models follows 12 layer (6 layer encoder + 6 layer decoder) architecture.

\subsection{Data Augmentation}

Apart from fine-tuning the model on actual test set, we also performed data augmentation and found a significant improvement over the results. For data augmentation, we performed 2 experiments. One by augmenting the 3X data to the actual dataset. Another by appending 5X data to the actual dataset. We found out that the performance of the model increased with increase in data augmentation amount.

\subsection{Training Configuration}

We used HuggingFace API and PyTorch to fine-tune the models. We used a learning rate of 2e-5. Maximum input and output sequence length as 1024 and 100 respectively. And fine-tuned for 5,7, 10 epochs for different experiments.

\section{Results}

This section gives a detailed overview of results containing all the experiments performed on the validation set. While table \ref{tab:val_eng}, table \ref{tab:val_hin} and table \ref{tab:val_guj} gives results of various experiments on validation dataset for English, Hindi and Gujarati respectively. Along with that table \ref{tab:test_eng}, table \ref{tab:test_hin} and table \ref{tab:test_guj} shows the final 3 test set results per language.

\subsection{Experiment Name}

This subsection defines the experiment name with their details which are mentioned in the below mentioned tables:

\subsubsection{English Experiments}

\begin{enumerate}
    \item \textbf{da\_en\_mt5}: mT5-small was finetuned in this approach along with data augmentation to 3 times of the actual english data.
    \item \textbf{da\_en\_ibart}: IndicBART was finetuned in this approach along with data augmentation to 3 times of the actual english data.
    \item \textbf{da5\_en\_ibart}: IndicBART was finetuned in this approach along with data augmentation to 5 times of the actual english data.
    \item \textbf{en\_ibart}: IndicBART was finetuned in this approach on the actual english dataset.
    \item \textbf{en\_mt5}: mt5-small was finetuned in this approach on the actual english dataset.
\end{enumerate}

\subsubsection{Hindi Experiments}
\begin{enumerate}
    \item \textbf{da5\_hi\_ibart}: IndicBART was finetuned in this approach along with data augmentation to 5 times of the actual hindi data.
    \item \textbf{da\_hi\_ibart}: IndicBART was finetuned in this approach along with data augmentation to 3 times of the actual hindi data.
    \item \textbf{da\_hi\_mt5}: mT5-small was finetuned in this approach along with data augmentation to 3 times of the actual hindi data.
    \item \textbf{hi\_ibart}: IndicBART was finetuned in this approach on the actual hindi dataset.
     \item \textbf{hi\_mt5}: mT5-small was finetuned in this approach on the actual hindi dataset.
\end{enumerate}

\subsubsection{Gujarati Experiments}

\begin{enumerate}
    \item \textbf{gu\_ibart}: IndicBART was finetuned in this approach on the actual gujarati dataset.
     \item \textbf{da\_gu\_ibart}: IndicBART was finetuned in this approach along with data augmentation to 3 times of the actual gujarati data.
     \item \textbf{da5\_gu\_ibart}: IndicBART was finetuned in this approach along with data augmentation to 5 times of the actual gujarati data.
      \item \textbf{gu\_mt5}: mT5-small was finetuned in this approach on the actual gujarati dataset.
\end{enumerate}

\subsection{Validation set results}

Below 3 tables shows results of our experiments on the validation set.


\begin{table*}[th]
  \caption{ROUGE F1 scores on English Validation set}
  \label{tab:val_eng}
  \begin{tabular}{ccccc}
    \toprule
    \textbf{Experiment} & \textbf{ROUGE-1} & \textbf{ROUGE-2} & \textbf{ROUGE-3} & \textbf{ROUGE-4} \\
    \midrule
    \multirow{1}{*}{da\_en\_mt5} & 0.54 & 0.43 & 0.41 & 0.40 \\
    \midrule
     \multirow{1}{*}{da\_en\_ibart} & 0.51 & 0.38 & 0.36 & 0.35 \\
    \midrule \\
     \multirow{1}{*}{da5\_en\_ibart} & 0.51 & 0.38 & 0.36 & 0.35 \\
    \midrule
     \multirow{1}{*}{en\_ibart} & 0.49 & 0.36 & 0.33 & 0.32 \\
     \midrule
     \multirow{1}{*}{en\_mt5} & 0.47 & 0.34 & 0.32 & 0.31 \\
  \bottomrule
\end{tabular}
\end{table*}


\begin{table*}[th]
  \caption{ROUGE F1 scores on Hindi Validation set}
  \label{tab:val_hin}
  \begin{tabular}{ccccc}
    \toprule
    \textbf{Experiment} & \textbf{ROUGE-1} & \textbf{ROUGE-2} & \textbf{ROUGE-3} & \textbf{ROUGE-4} \\
    \midrule
    \multirow{1}{*}{da5\_hi\_ibart} & 0.6104 & 0.515 & 0.488 & 0.475 \\
    \midrule
     \multirow{1}{*}{da\_hi\_ibart} & 0.604 & 0.508 & 0.482 & 0.470 \\
    \midrule \\
     \multirow{1}{*}{da\_hi\_mt5} & 0.595 & 0.49 & 0.473 & 0.46 \\
    \midrule
     \multirow{1}{*}{hi\_ibart} & 0.594 & 0.497 & 0.471 & 0.458 \\
     \midrule
     \multirow{1}{*}{hi\_mt5} & 0.54 & 0.438 & 0.412 & 0.398 \\
  \bottomrule
\end{tabular}
\end{table*}

\begin{table*}[th]
  \caption{ROUGE F1 scores on Gujarati Validation set}
  \label{tab:val_guj}
  \begin{tabular}{ccccc}
    \toprule
    \textbf{Experiment} & \textbf{ROUGE-1} & \textbf{ROUGE-2} & \textbf{ROUGE-3} & \textbf{ROUGE-4} \\
    \midrule
    \multirow{1}{*}{gu\_ibart} & 0.246 & 0.146 & 0.118 & 0.105 \\
    \midrule
     \multirow{1}{*}{da\_gu\_ibart} & 0.239 & 0.144 & 0.118 & 0.105 \\
    \midrule \\
     \multirow{1}{*}{da5\_gu\_ibart} & 0.235 & 0.137 & 0.11 & 0.096 \\
     \midrule
     \multirow{1}{*}{gu\_mt5} & 0.206 & 0.114 & 0.09 & 0.079 \\
  \bottomrule
\end{tabular}
\end{table*}

\subsection{Test set results}

The below 3 tables shows the results of top 3 experiments per language on official test set.

\begin{table*}[th]
  \caption{ROUGE F1 scores on English Test set}
  \label{tab:test_eng}
  \begin{tabular}{ccccc}
    \toprule
    \textbf{Experiment} & \textbf{ROUGE-1} & \textbf{ROUGE-2} & \textbf{ROUGE-3} & \textbf{ROUGE-4} \\
    \midrule
     \multirow{1}{*}{da5\_en\_ibart} & 0.521 & 0.401 & 0.378 & 0.369 \\
    \midrule \\
     \multirow{1}{*}{da\_en\_ibart} & 0.512 & 0.389 & 0.366 & 0.358 \\
    \midrule
     \multirow{1}{*}{en\_ibart} & 0.493 & 0.367 & 0.344 & 0.336 \\
   \bottomrule
\end{tabular}
\end{table*}

\begin{table*}[th]
  \caption{ROUGE F1 scores on Hindi Test set}
  \label{tab:test_hin}
  \begin{tabular}{ccccc}
    \toprule
    \textbf{Experiment} & \textbf{ROUGE-1} & \textbf{ROUGE-2} & \textbf{ROUGE-3} & \textbf{ROUGE-4} \\
    \midrule
     \multirow{1}{*}{da5\_hi\_ibart} & 0.592 & 0.491 & 0.464 & 0.451 \\
    \midrule \\
     \multirow{1}{*}{da\_hi\_ibart} & 0.586 & 0.485 & 0.458 & 0.445 \\
    \midrule
     \multirow{1}{*}{hi\_mt5} & 0.544 & 0.438 & 0.41 & 0.397 \\
   \bottomrule
\end{tabular}
\end{table*}

\begin{table*}[th]
  \caption{ROUGE F1 scores on Gujarati Test set}
  \label{tab:test_guj}
  \begin{tabular}{ccccc}
    \toprule
    \textbf{Experiment} & \textbf{ROUGE-1} & \textbf{ROUGE-2} & \textbf{ROUGE-3} & \textbf{ROUGE-4} \\
    \midrule
     \multirow{1}{*}{da5\_gu\_ibart} & 0.242 & 0.146 & 0.119 & 0.106 \\
    \midrule \\
     \multirow{1}{*}{da\_gu\_ibart} & 0.241 & 0.145 & 0.120 & 0.107 \\
    \midrule
     \multirow{1}{*}{gu\_mt5} & 0.203 & 0.115 & 0.094 & 0.084 \\
   \bottomrule
\end{tabular}
\end{table*}

\subsection{Analysis}

From the above results, we can say that data augmentation is a useful step as it has shown significant improvement of results over other experiments. Also, on comparing IndicBART and mT5, we can say that IndicBART performed better in most of the cases than mT5 for the summarization task. Further improvement can be made by using larger models like mbart-large or mt5-base/mt5-large models.

\section{Conclusion}

In this work, we presented our work for performing summarizing indian languages as part of the Forum for Information Retrieval Evaluation, 2022 shared task. We perform various experiments with multilingual transformer based models like IndicBART and mT5-small and acheived significant results. For Hindi and Gujarati languages, we stood at $2^{nd}$ place. While for English language, we stood at $4^{th}$ place. Due to computational constraints, we were not able to use larger models like mbart-large and mt5-base which could have performed even better. We hope this work will help future research in this direction.

\bibliography{main}

\appendix

\end{document}